\documentclass{article}
\usepackage{spconf,amsmath,graphicx}

\usepackage{times}  
\usepackage{helvet}  
\usepackage{courier}  
\usepackage[hyphens]{url}  
\usepackage{caption} 
\usepackage{algorithm}
\usepackage{algorithmic}

\usepackage{placeins}
\usepackage{multirow}
\usepackage{booktabs}       

\usepackage{amsfonts}       
\usepackage{nicefrac}       
\usepackage{microtype}      
\usepackage{xcolor}         
\usepackage{epsfig}
\usepackage{amssymb}
\usepackage{subfigure}
\usepackage{listings}

\title{TFDMNet: A Novel Network Structure Combines the Time Domain and Frequency Domain Features}
\name{Hengyue Pan, Yixin Chen, Zhiliang Tian, Peng Qiao, Linbo Qiao, Dongsheng Li}
\address{School of Computer \\
	National University of Defense Technology\\
	109 Deya Road, Changsha, China 410073}

\begin{document}
%
\maketitle
\begin{abstract}
Convolutional neural network (CNN) has achieved impressive success in computer vision during the past few decades. The image convolution operation helps CNNs to get good performance on image-related tasks. However, it also has high computation complexity and hard to be parallelized. This paper proposes a novel Element-wise Multiplication Layer (EML) to replace convolution layers, which can be trained in the frequency domain. Theoretical analyses show that EMLs lower the computation complexity and easier to be parallelized. Moreover, we introduce a Weight Fixation mechanism to alleviate the problem of over-fitting, and analyze the working behavior of Batch Normalization and Dropout in the frequency domain. To get the balance between the computation complexity and memory usage, we propose a new network structure, namely Time-Frequency Domain Mixture Network (TFDMNet), which combines the advantages of both convolution layers and EMLs. Experimental results imply that TFDMNet achieves good performance on MNIST, CIFAR-10 and ImageNet databases with less number of operations comparing with corresponding CNNs. 
\end{abstract}
\section{Introduction}
\label{sec:intro}
In the past few decades, CNN \cite{lecun1995convolutional} has played an essential role in computer vision. Even though ViT \cite{dosovitskiy2021an} shows excellent performance on vision tasks, CNN is still an irreplaceable tool due to its lower number of free-parameters and less requirements for training data. The core of CNN is the image convolution operation, which has high computation complexity. Based on theorems of signal processing, image convolution can be replaced by the straightforward element-wise multiplication via converting input data and convolution filters into the frequency domain using Discrete Fourier Transform (DFT). In this way, we may simplify the forward and backward calculations of convolution layers and make them easier to be parallelized. 

There are many efforts to implement convolution layers in the frequency domain. \cite{LeCunDFTDNN2014} is an early research that considered to do convolution operation in the frequency domain. The method achieved good efficiency in the inference stage, but it could not learn convolution filters directly in the frequency domain. \cite{Vasilache2015FastCN}, \cite{FastCNN}, and \cite{FALCON} also mainly focused on the inference stage. Those researches implied that transferring well-trained CNNs into the frequency domain may obviously speed-up the inference computation. \cite{IEEEAccess} proposed a neural network training framework in the frequency domain and has good performance on Meta-Gram and ImageNet datasets. However, the Fourier domain training algorithm of \cite{IEEEAccess} follows \cite{LeCunDFTDNN2014}, which means it still needs to move weights and features between the time domain and the frequency domain frequently. 

In \cite{DFTPooling}, DFT was applied on pooling layers of CNNs. \cite{WATANABE2021107851} proposed 2SReLU layer that can work in the frequency domain. \cite{FCNN} proposed a Fourier Convolution Neural Network (FCNN), which can be trained entirely in the frequency domain. 

One drawback of the element-wise multiplication is the high memory usage, especially for the large-scale databases and network structures. To get a balance between the number of computation operation and memory usage, in this paper, we propose a novel computation model, namely Time-Frequency Domain Mixture Network (TFDMNet), which combines the merits of both image convolutions and element-wise multiplications. The main contributions of this paper include:

(1) We propose a Element-wise Multiplication Layer (EML) with Weight Fixation that can be trained directly in the frequency domain to replace the image convolution operation of CNNs for a lower computation complexity and higher parallelizability. 

(2) We propose a novel model, namely TFDMNet, to combine the merits of convolution layers and EMLs.

(3) We implement Batch Normalization and Dropout in the frequency domain to improve the performance of TFDMNet, and design a two-branches structure for TFDMNet to make it work with complex inputs.

\section{Mathematical Basis}
\label{sec:Math}

%
%
%
%
%
%

In the field of signal processing, {\bfseries Cross-Correlation Theorem} is one of the basic calculation rule. Assuming that ${\bf u}$ and ${\bf v}$ are two casual signals in the time domain, $\ast$ the image convolution operation (in fact the cross-correlation operation in signal processing), and $\mathcal{F}(\cdot)$ the Discrete Fourier transform (DFT), then we have:

\begin{equation}
	\label{eq:cct}
	\mathcal{F}(R({\bf u}, {\bf v})) = \mathcal{F}^*({\bf u}) \cdot \mathcal{F}({\bf v})
\end{equation}

where $R({\bf u}, {\bf v})$ is the cross-correlation between ${\bf u}$ and ${\bf v}$, and $\mathcal{F}^*({\bf u})$ is the conjugate complex number of $\mathcal{F}({\bf u})$. 

In convolution layers of CNNs, the so-called image convolution operation is in fact the cross-correlation between inputs and convolution filters. Therefore, Eq.~\ref{eq:cct} serves as an important basis of our research since it builds a relationship between the time domain and frequency domain.

\section{Method}
\label{sec:method}

In this section, we firstly introduce necessary network layers in the frequency domain, then propose the implementation method of the TFDMNet.

\subsection{Important Layers}

\subsubsection{Element-wise Multiplication Layer (EML)}

Based on Section~\ref{sec:Math}, the image convolution operation in the time domain can be replaced by element-wise multiplication in the frequency domain. Therefore, we design the EML as the most important part of the TFDMNet. 

Assuming that we have a convolution layer $L$, which has $H_1 \times H_2 \times C_{in} $ sized input feature map ${\bf I}_L$ and $H_1 \times H_2 \times C_{out} $ sized output feature map ${\bf O}_L$. Notice that here we only consider the situation that ${\bf I}_L$ and ${\bf O}_L$ have the same $H_1$ and $H_2$. The convolution filter ${\bf W}$ of $L$ has the size of $K \times K \times C_{in} \times C_{out}$, where $K \le H_1$ and $K \le H_2$. Thus the forward process of $L$ is :

\begin{equation}
	\label{convforward}
	{\bf O}_L = {\bf I}_L \ast {\bf W}
\end{equation}

where $\ast$ is the image convolution operation. According to Eq.~\ref{eq:cct}, the corresponding operation of Eq.~\ref{convforward} in the frequency domain is:

\begin{equation}
	\label{eq:convforwardfreq}
	\mathcal{F}({\bf O}_L) = \mathcal{F}^*({\bf I}_L) \cdot \mathcal{F}({\bf W}_p)
\end{equation}

Notice that we should firstly do zero-padding on ${\bf W}$ to generate the padded filter ${\bf W}_p$ to guarantee that it has the same height and width as ${\bf I}_L$. Moreover, we should expand $\mathcal{F}^*({\bf I}_L)$ to $H_1 \times H_2 \times C_{in} \times C_{out}$ by simply copy it for $C_{out}$ times.  At the end of the calculation, we sum over the third dimension of $\mathcal{F}({\bf O}_L)$ to make it has the size of $H_1 \times H_2 \times C_{out} $.

We can easily find that the computation complexity of one convolution layer is $O(K^2 \times H_1 \times H_2 \times C_{in} \times C_{out})$, while for one EML it is reduced to $O(H_1 \times H_2 \times C_{in} \times C_{out})$.

Eq.~\ref{eq:convforwardfreq} is the forward calculation of EMLs, and it is very easy to derive the gradient of the layer: $\frac{\partial \mathcal{F}({\bf O}_L)}{\partial \mathcal{F}({\bf W}_p)} = \mathcal{F}^*({\bf I}_L)$. Here we can learn that the gradient calculation of EMLs is much easier than regular convolution layers. Moreover, it is obvious that EMLs are easier to be parallelized than convolution layers.

It is easy to know that $\mathcal{F}({\bf W}_p)$ has much more free parameters than ${\bf W}$, which may result in over-fitting. To fix this problem, we introduce a {\bfseries Weight Fixation} mechanism during the training process. Specifically, after each weight updating, we transfer $\mathcal{F}({\bf W}_p)$ back to the time domain (still denoted by ${\bf W}_p$), and perform an element-wise multiplication between ${\bf W}_p$ and a Weight Fixation matrix ${\bf V}$, where ${\bf V}$ is a $0-1$ matrix and only upper-left $K \times K$ elements are set to $1$. Thus Eq.~\ref{eq:convforwardfreq} becomes: $\mathcal{F}({\bf O}_L) = \mathcal{F}^*({\bf I}_L) \cdot \mathcal{F}({\bf W}_p \cdot {\bf V})$. In this way, we introduce a restrict to $\mathcal{F}({\bf W}_p)$ to guarantee that it has the same number of free parameters as the corresponding ${\bf W}$.

The computation complexity of DFT for Weight Fixtation is $O(H_1 \times H_2 \times C_{in} \times C_{out} \times log(H_1 \times H_2))$. When $H_1 = H_2 = H$, the computation complexity of DFT reduced to $O(H^2 \times C_{in} \times C_{out} \times log(H))$. Even though the proposed Weight Fixation mechanism slows down the training process, it can obviously improve network performance. 

\subsubsection{Batch Normalization}

Batch Normalization \cite{ioffe15} is a widely-used regularization method in deep learning. Assuming that ${\bf B} = \{{\bf u}_1, {\bf u}_2, ..., {\bf u}_S\}$ is a training mini-batch with batch size $S$, the basic procedure of Batch Normalization can be divided into two steps:

1. Normalization: we should firstly calculate the mean $\mu_{\bf B}$ and variance $\sigma_{\bf B}$ over the mini-batch ${\bf B}$, then normalize training samples in ${\bf B}$: $\hat{{\bf u}}_i = \frac{{\bf u}_i - \mu_{\bf B}}{\sqrt{\sigma_{\bf B}^2 + \epsilon}}, i = 1, ..., S$, where $\epsilon$ is a small enough constant to prevent zero-denominator. 

2. Scale and shift:  in this step two learnable parameters $\gamma$ and $\beta$ are introduced to perform the Batch Normalization Transform on $\hat{x_i}$: $BNT_{\gamma, \beta}({\bf u}_i) = \gamma \hat{{\bf u}_i} + \beta, i = 1, ..., S$

Based on the definition of Discrete Fourier Transform ($\mathcal{F}_{real}(u,v) = \sum_{x=0}^{M-1}\sum_{y=0}^{N-1} {\bf u}_i (x,y) \cos(2\pi(\frac{ux}{M}+\frac{vy}{N}))$ and $\mathcal{F}_{imag}(u,v) = -\sum_{x=0}^{M-1}\sum_{y=0}^{N-1} {\bf u}_i (x,y) \sin(2\pi(\frac{ux}{M}+\frac{vy}{N})) $), we transfer above time domain data ${\bf u}_i$ to the frequency domain, and assuming that $\mathcal{F}_{real}(u,v)$ and $\mathcal{F}_{imag}(u,v)$ are the real part and imaginary part of the corresponding training data in the frequency domain respectively (where $u, v$ are the coordinates). If we perform Batch Normalization on the time domain features, it is easy to get the frequency domain counterparts $\mathcal{F}_{i,real}^{BNT}(u,v)$, $\mathcal{F}_{i,imag}^{BNT}(u,v)$, $\mu_{\mathcal{F}, real}$, $\mu_{\mathcal{F}, imag}$, $\sigma_{\mathcal{F}, real}$ and $\sigma_{\mathcal{F}, imag}$. Finally we have:
\begin{normalsize}
\begin{equation}
	\label{eq:DFTreal3}
	\begin{array}{cll}
		& &\mathcal{F}_{i,real}^{BNT}(u,v) \\
		& & \\
		& = &\frac{\gamma}{\sqrt{C_{real}}} \frac{\mathcal{F}_{i,real}(u,v) - \mu_{\mathcal{F}, real}}{\sqrt{\sigma_{\mathcal{F}, real}^2 + \epsilon}} + \beta_{real}
	\end{array}
\end{equation}
\end{normalsize}
and 
\begin{normalsize}
\begin{equation}
	\label{eq:DFTimag2}
	\begin{array}{cll}
		& &\mathcal{F}_{i,imag}^{BNT}(u,v) \\
		& & \\
		& = &\frac{\gamma}{\sqrt{C_{imag}}} \frac{\mathcal{F}_{i,imag}(u,v) - \mu_{\mathcal{F}, imag}}{\sqrt{\sigma_{\mathcal{F}, imag}^2 + \epsilon}} + \beta_{imag}
	\end{array}
\end{equation}
\end{normalsize}
where $i = 1, ..., S$, and $C_{real} = \frac{\sigma_{\bf B}^2}{\sigma_{\mathcal{F}, real}^2}$ and $C_{imag} = \frac{\sigma_{\bf B}^2}{\sigma_{\mathcal{F}, imag}^2}$ are constants. Thus $\frac{\gamma}{\sqrt{C_{real}}}$ and $\frac{\gamma}{\sqrt{C_{imag}}}$ can be viewed as learnable parameters.

Based on the analyses above, we can learn that the implementation of Batch Normalization in the frequency domain has exactly the same form as the time domain. Thus in practice, we can directly do Batch Normalization on the real part and imaginary part of complex features in each mini-batch respectively. 

\subsubsection{Approximated Dropout}

Dropout \cite{srivastava2014dropout} is another widely-used regularization method for deep neural networks. Specifically, every neuron of the network may be dropped with probability $p$ during the training process. Assuming that ${\bf u}_i (x,y)$ is one of the neurons of the time domain feature. Then the Dropout can be written as ${\bf u}_{id}(x, y) = r {\bf u}_i (x, y)$, where $r$ follows the Bernoulli distribution with probability $1-p$ (which means the Dropout rate is $p$).

Based on definition of DFT and Dropout, we propose an approximation method to implement Dropout in the frequency domain. The foundation of our approximation method is the observation that performing Dropout on ${\bf u}_i (x, y)$ equals to randomly shrink or amplify $\mathcal{F}_{real}(u,v)$ and $\mathcal{F}_{imag}(u,v)$, because during the DFT calculation the values of both $\cos(2\pi(\frac{ux}{M}+\frac{vy}{N}))$ and $\sin(2\pi(\frac{ux}{M}+\frac{vy}{N}))$ have the same probability that within the ranges of $[-1, 0]$ and $(0, 1]$. Therefore, we have:

\begin{equation}
	\label{eq:DFTdroppedappr}
	\begin{split}
		& \mathcal{F}_{d,real}(u,v) \simeq r_{real}(u,v) \mathcal{F}_{real}(u,v) \\
		& \mathcal{F}_{d,imag}(u,v) \simeq r_{imag}(u,v) \mathcal{F}_{imag}(u,v)
	\end{split}
\end{equation}

Where $\mathcal{F}_{d}$ is the dropped data in the frequency domain. Based on the analyses above, we make an assumption that $r_{real}(u,v)$ and $r_{imag}(u,v)$ should not have large deviation from $1$, since all elements of the time domain features have the same probability to be dropped. It is easy to learn that $r_{real}(u,v)$ and $r_{imag}(u,v)$ have higher probability to lay between $1-p$ and $1+p$. Therefore, we make $r_{real}(u,v)$ and $r_{imag}(u,v)$ obey a normal distribution $\mathcal{N}(1, p/2)$, and in this case the probability that $r_{real}(u,v)$ and $r_{imag}(u,v)$ lay between $1-p$ and $1+p$ is about $95.4\%$. Notice that we ignore the relationship between $\cos(2\pi(\frac{ux}{M}+\frac{vy}{N}))$ and $\sin(2\pi(\frac{ux}{M}+\frac{vy}{N}))$ to simplify the calculation, thus $r_{real}(u,v)$ and $r_{imag}(u,v)$ are independent with each other. In practice we set $p=0.5$ for all layers, and the proposed approximated Dropout shows good performance in the frequency domain.

\subsubsection{Max Pooling}

Max Pooling is used for down-sampling in CNNs. Unfortunately, Max Pooling cannot work in the frequency domain since complex numbers cannot compare with each other. Therefore, we firstly transfer the input feature maps back to the time domain, then perform Max Pooling and transfer the results back to the frequency domain. The computation complexities of the DFT and iDFT operation here are $O(H_1 \times H_2 \times C_{in} \times log(H_1 \times H_2))$ and $O(H_1^{'} \times H_2^{'} \times C_{out} \times log(H_1^{'} \times H_2^{'}))$, respectively. Where $H_1^{'}$ and $H_2^{'}$ are the size of down-sampled features.

\subsection{The Implementation of Complex Layers}

Based on the analysis above, it is easy to know that we can process real parts and imaginary parts of input features separately for most kinds of layers. Inspired by \cite{Guberman16ComplexCNN} and \cite{trabelsi2018deep}, we design a two-branches structure to integrate network layers in the frequency domain. Specifically, we use one branch for the real part and the other for the imaginary part.

\subsection{Time-Frequency Domain Mixture Network}

One important drawback of the above-mentioned EMLs is the high memory usage, since the small-scale convolution filters should be padded to much larger weight matrices in the frequency domain. Therefore, we propose a Time-Frequency Domain Mixture Network (TFDMNet) to meet a suitable balance between memory usage and number of operations. For the shallow layers that contain larger-sized feature maps, we let them work in the time-domain to reduce the number of parameters, while for the deeper layers with small-sized feature maps, we transfer them to the frequency domain to reduce the number of operations. 

At the end of TFDMNet, we flatten the real part and imaginary part of complex feature maps, then feed them into one or more fully connected layers. Finally, we concatenate the real and imaginary feature vectors, and use one fully connected layer to generate classification results. 

\section{Experimental Results} 
\label{sec:experiments}

\subsection{Databases and Computation Platform}

In this paper, we apply three widely used databases to evaluate the proposed TFDMNet, i.e., MNIST \cite{lecun1998gradient}, CIFAR-10  \cite{krizhevsky2014cifar} and ImageNet \cite{krizhevsky2012imagenet}. We implement the proposed TFDMNet on Tensorflow 2.5 \cite{tensorflow2015-whitepaper}, and our computation platform includes Intel Xeon 4108 CPU, 256 GB memory, and eight Tesla A40 GPUs. Our codes are available on https://github.com/mowangphy88/TFDMNet.

\subsection{MNIST Experiments}

The baseline CNN structure of MNIST experiments is based on LeNet-5 \cite{LeNet}, and we add batch normalization and dropout layers into it. Since MNIST is a small scale problem, we set all layers of the TFDMNet work in the frequency domain. During the training process of TFDMNet, we use RMSProp as the optimizer. The batch-size is set to $100$, and the network should be trained for $800$ epochs. Notice that we do not apply any data augmentation methods during the training process.



We select several state-of-the-art frequency domain neural network methods as baselines to compare with TFDMNet. Moreover, to show the importance of the proposed methods, we also include ablation studies. The experimental results of MNIST are included in Table~\ref{table-MNIST}, and we can learn that Weight Fixation, Batch Normalization, and Approximated Dropout work well in the frequency domain. 

\begin{table}[t]
	\vskip 0.1in
	\begin{center}
		\begin{small}
			\begin{sc}
				\caption{Experimental results on MNIST. (WF: Weight Fixation, BN: Batch Normalization, DO: Dropout.)}
				\label{table-MNIST}
				\begin{tabular}{c|c|c}
					\toprule
					Methods                             &   \# of Ops   & Test Err     \\
					\midrule
					Modified LeNet-5                    &  $\approx$ 692K      & 0.72\%        \\
					\midrule
					2SReLU \cite{WATANABE2021107851}     &    N/A                & 3.08\%        \\
					FCNN \cite{FCNN}                     &    N/A                & $\approx$ 3\%  \\
					\midrule
					TFDMNet - WF - BN - DO           &    N/A                & 1.11\%        \\
					TFDMNet - BN - DO               &    N/A                  & 0.99\%         \\
					TFDMNet - Dropout                     &    N/A                 & 0.91\%         \\
					TFDMNet (all freq)         &   $\approx$ 481K &       0.63\%         \\
					\bottomrule
				\end{tabular}
			\end{sc}
		\end{small}
	\end{center}
	\vskip -0.1in
\end{table}

\begin{table}[ht]
	\vskip 0.1in
	\begin{center}
		\begin{small}
			\begin{sc}
				\caption{Experimental results on CIFAR-10. }
				\label{table-CIFAR}
				\begin{tabular}{c|c|c}
					\toprule
					Methods                              & \# of Ops      & Test Err     \\
					\midrule
					Small CNN                             &  $\approx$ 68.10M      & 21.07\%    \\
					Large CNN                             &  $\approx$ 275.45M    & 11.30\%    \\
					\midrule
					FCNN \cite{FCNN}                     &    N/A                  & $\approx$ 73\%  \\
					\midrule
					Small TFDMNet(all freq)     &   $\approx$ 20.65M   & 22.40\%     \\
					Large TFDMNet(all freq)      &   $\approx$ 84.62M   & 21.63\%       \\
					Large TFDMNet(mixture)            &   $\approx$ 181.47M    & 11.20\%       \\
					\bottomrule
				\end{tabular}
			\end{sc}
		\end{small}
	\end{center}
	\vskip -0.1in
\end{table}

\subsection{CIFAR-10 Experiments}

The baseline CNN structure of CIFAR-10 experiments is based-on VGG-16 \cite{simonyan2014vgg} framework. We consider two scales of networks:

(1) Small scale networks: each block contains one convolution layer or EML, and the network ends with one fully-connected layer with $512$ neurons. Notice that all layers of the small scale TFDMNet work in the frequency domain.

(2) Large scale networks: the number of layers in each block is set to $2, 2, 2, 3, 3$, and the network ends with one fully-connected layer with $512$ neurons (For the TFDMNet, the first 3 blocks work in the time domain, while the last two blocks and all fully-connnected layers work in the frequency domain).

%

The training process of TFDMNet with CIFAR-10 database shares similar optimizer and initialization method with MNIST, and no data augmentation methods are applied.



From experiments of CIFAR-10 we can learn that the Small TFDMNet has comparable performance to the corresponding Small CNN due to the shallow structure. For the Large TFDMNet, if we make all layers work in the frequency domain, the performance gap with the corresponding Large CNN is relatively large. Fortunately, by using mixture model that combines benefits of convolution layers and EMLs, we successfully fix this problem and achieve good classification performance with less number of operations.

\subsection{ImageNet Experiments}

The baseline CNN structure of ImageNet experiments is AlexNet \cite{NIPS2012_AlexNet}, which contains 5 convolution layers and 3 pooling layers. The network ends with two fully-connected layers with 4096 neuros. We build the corresponding TFDMNets based on AlexNet. We set the first layer works in the time domain, while the rest of layers work in the frequency domain. We use SGD to train the networks for $120$ epochs with the batch-size $256$. Table~\ref{table-ImageNet} provides experimental results on ImageNet, and we can learn that TFDMNet obviously reduce the computation complexity, and achieves comparable classification accuracy with its CNN counterpart.

\begin{table}[ht]
	\vskip 0.1in
	\begin{center}
		\begin{small}
			\begin{sc}
				\caption{Experimental results on ImageNet.}
				\label{table-ImageNet}
				\begin{tabular}{c|c|c}
					\toprule
					Methods                            & \# of Ops        & Top-1 Error     \\
					\midrule
					AlexNet                            &  $\approx$ 1.48B    & 41.77\%    \\
					\midrule
					TFDMNet                &   $\approx$ 0.94B    & 44.60\%    \\
					\bottomrule
				\end{tabular}
			\end{sc}
		\end{small}
	\end{center}
	\vskip -0.1in
\end{table}.



\section{Conclusion}
\label{sec:concl}

In this paper, we propose TFDMNet, which is a mixture model of the time and frequency domain layers to achieve a balance between the computation complexity and memory usage. Based on the Cross-Correlation Theorem, we design EMLs to replace convolution layers, which has lower computation complexity and easier to be parallelized. Moreover, we theoretically analyze the working behaviour of Batch Normalization and Dropout in the frequency domain, and introduce their counterparts into TFDMNet. Also, to deal with complex inputs brought by DFT, we design a two-branches network structure for the frequency domain parts of TFDMNet. Experimental results show that TFDMNets achieve good performance on MNIST, CIFAR-10 and ImageNet databases with lower computation complexity. 


\bibliographystyle{IEEEbib}
\bibliography{refs}

\end{document}